%% file: main.tex
\documentclass[11pt,a4paper]{article}
\usepackage[hyperref]{acl2019}
\usepackage{times}
\usepackage{latexsym}
\usepackage{subfig}
\usepackage{bm}
\usepackage{amsmath}
\usepackage{amsfonts}
\usepackage{booktabs}
\usepackage{enumitem}
\usepackage{colortbl}
\usepackage{graphicx}
\usepackage{bbm}
\usepackage{multirow}
\usepackage{pbox}
\usepackage{arydshln}
\usepackage[T1]{fontenc}
\usepackage{soul}
\DeclareCaptionLabelFormat{andtable}{#1~#2  \&  \tablename~\thetable}
\usepackage{url}
\usepackage{tikz}
\newcommand*\circled[1]{\tikz[baseline=(char.base)]{
            \node[shape=circle,draw,inner sep=0.4pt] (char) {#1};}}
\aclfinalcopy

\title{Argument Generation with Retrieval, Planning, and Realization}

\author{Xinyu Hua,  Zhe Hu, {\rm and} Lu Wang\\
  Khoury College of Computer Sciences\\ Northeastern University\\ Boston, MA 02115 \\
  {\tt \{hua.x, hu.zhe\}@husky.neu.edu}, {\tt luwang@ccs.neu.edu} 
 }

\date{}

\begin{document}
\maketitle

\input{000abstract.tex}

\section{Introduction}
\label{sec:intro}
\input{010intro.tex}

\section{Related Work}
\label{sec:related}
\input{020related.tex}

\section{Overview of \textsc{CANDELA}}
\label{sec:overview}
\input{030overview.tex}

\section{Argument Retrieval}
\label{sec:retrieval}
\input{040retrieval.tex}

\section{Argument Generation}
\label{sec:model}
\input{050model.tex}

\section{Experimental Setups}
\label{sec:experiment}
\input{060experiment.tex}
\section{Results and Analysis}
\label{sec:analysis}
\input{070analysis.tex}

\section{Conclusion}
\label{sec:conclusion}
\input{080conclusion.tex}

\section*{Acknowledgements}
This research is supported in part by National Science Foundation through Grants IIS-1566382 and IIS-1813341. We thank Varun Raval for helping with data processing and search engine indexing. 
We are grateful to the three anonymous reviewers for their constructive suggestions. 

\bibliography{acl2019}
\bibliographystyle{acl_natbib}

\appendix
\section{Appendices}
\input{090appendix.tex}
\input{100samples.tex}

\end{document}

%% file: 000abstract.tex
\begin{abstract}
Automatic argument generation is an appealing but challenging task. 
In this paper, we study the specific problem of counter-argument generation, and present a novel framework, \textsc{CANDELA}. It consists of a powerful retrieval system and a novel two-step generation model, where a {\bf text planning decoder} first decides on the main talking points and a proper language style for each sentence, then a {\bf content realization decoder} reflects the decisions and constructs an informative paragraph-level argument. 
Furthermore, our generation model is empowered by a retrieval system indexed with $12$ million articles collected from Wikipedia and popular English news media, which provides access to high-quality content with diversity. 
Automatic evaluation on a large-scale dataset collected from Reddit shows that our model yields significantly higher BLEU, ROUGE, and METEOR scores than the state-of-the-art and non-trivial comparisons. Human evaluation further indicates that our system arguments are more appropriate for refutation and richer in content. 
\end{abstract}

%% file: 010intro.tex
Counter-argument generation aims to produce arguments of a different stance, in order to refute the given proposition on a controversial issue~\cite{toulmin1958use,damer2012attacking}. 
A system that automatically constructs counter-arguments can effectively present alternative perspectives along with associated evidence and reasoning, and thus facilitate a more comprehensive understanding of complicated problems when controversy arises.

\begin{figure}[t]
    \centering
    \includegraphics[width=78mm]{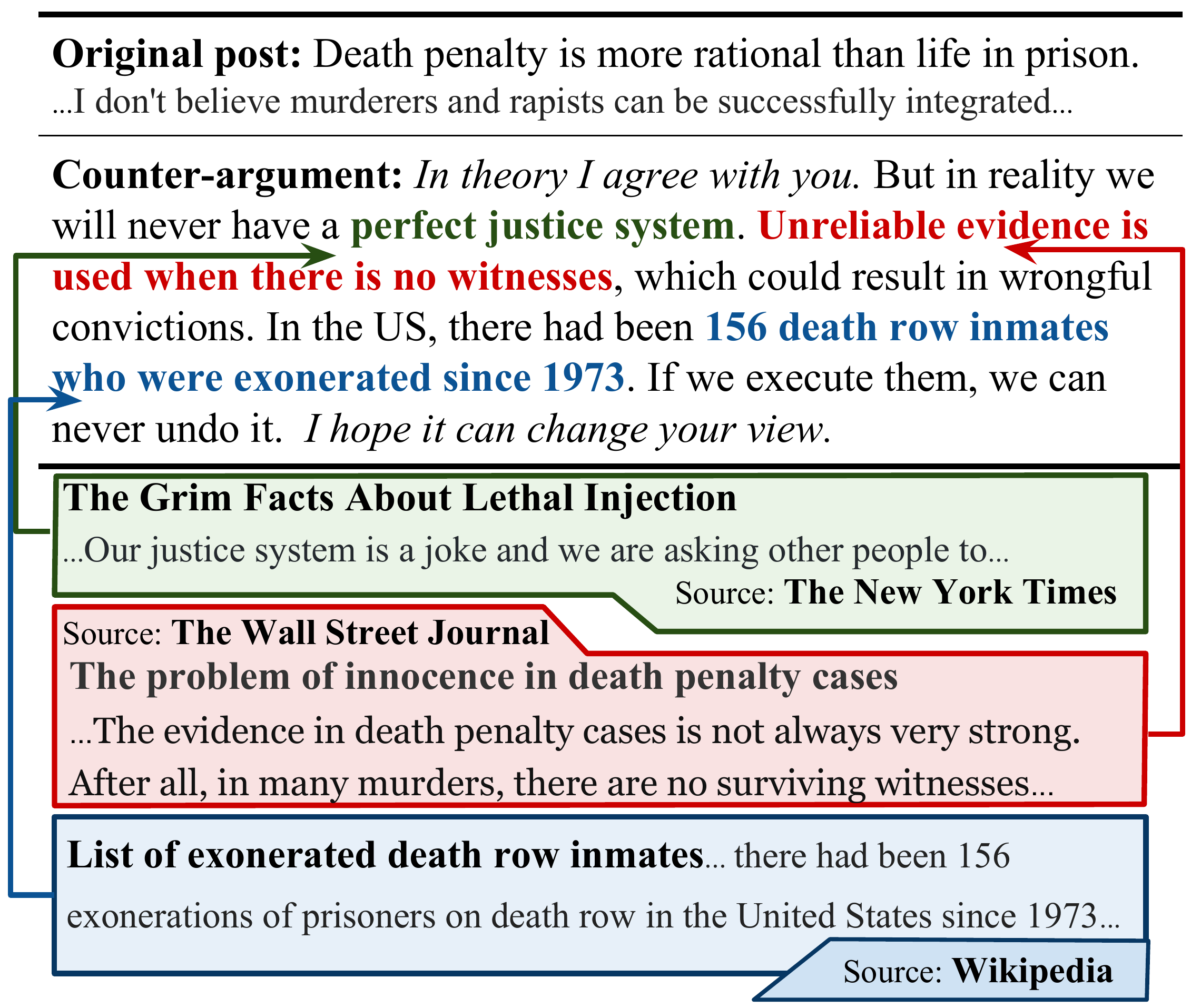}
    \caption{\fontsize{10}{12}\selectfont 
    Sample counter-argument for a pro-death penalty statement from Reddit \texttt{/r/ChangeMyView}. The argument consists of a sequence of propositions, by synthesizing opinions and facts from diverse sources. Sentences in italics contain stylistic languages for argumentation purpose.
    }
    \label{fig:motivating_example}
\end{figure}

Nevertheless, constructing persuasive arguments is a challenging task, as it requires an appropriate combination of credible evidence, rigorous logical reasoning, and sometimes emotional appeal~\cite{walton2008argumentation,P17-2039,Q17-1016}. A sample counter-argument for a pro-death penalty post is shown in Figure~\ref{fig:motivating_example}. As can be seen, a sequence of talking points on the ``imperfect justice system'' are presented: it starts with the fundamental concept, then follows up with more specific evaluative claim and supporting fact. 
Although retrieval-based methods have been investigated to construct counter-arguments~\cite{sato-EtAl:2015:ACL-IJCNLP-2015-System-Demonstrations,reisert-EtAl:2015:ARG-MINING}, they typically produce a collection of sentences from disparate sources, thus fall short of coherence and conciseness. 
Moreover, human always deploy {\it stylistic languages} with specific argumentative functions to promote persuasiveness, such as making a concessive move (e.g., ``{\it In theory I agree with you}"). 
This further requires the generation system to have better control of the languages style.

Our goal is to design {\it a counter-argument generation system to address the above challenges and produce paragraph-level arguments with rich-yet-coherent content}. 
To this end, we present \textsc{\bf CANDELA}---a novel framework to generate \underline{C}ounter-\underline{A}rguments with two-step \underline{N}eural \underline{D}ecoders and \underline{E}xterna\underline{L} knowledge \underline{A}ugmentation.\footnote{Code and data are available at \url{https://xinyuhua.github.io/Resources/acl19/}.} 
Concretely, \textsc{CANDELA} has three major distinct features: 

First, it is equipped with two decoders: one for \textbf{text planning}---selecting talking points to cover for {\it each sentence} to be generated, the other for \textbf{content realization}---producing a fluent argument to reflect decisions made by the text planner. This enables our model to produce longer arguments with richer information. 

Furthermore, multiple objectives are designed for our text planning decoder to both handle content selection and ordering, and select a proper argumentative discourse function of a desired language style for each sentence generation. 

Lastly, the input to our argument generation model is augmented with keyphrases and passages retrieved from a large-scale search engine, 
which indexes $12$ million articles from Wikipedia and four popular English news media of varying ideological leanings. This ensures access to reliable evidence, high-quality reasoning, and diverse opinions from different sources, as opposed to recent work that mostly considers a single origin, such as Wikipedia~\cite{rinott-EtAl:2015:EMNLP} or online debate portals~\cite{P18-1023}. 

We experiment with argument and counter-argument pairs collected from the Reddit \texttt{/r/ChangeMyView} group.  
Automatic evaluation shows that the proposed model significantly outperforms our prior argument generation system~\cite{P18-1021} and other non-trivial comparisons. 
Human evaluation further suggests that our model produces more appropriate counter-arguments with richer content than other automatic systems, while maintaining a fluency level comparable to human-constructed arguments.

%% file: 020related.tex
To date, the majority of the work on automatic argument generation leads to rule-based models, e.g., designing operators that reflect strategies from argumentation theory~\cite{reed1996architecture,carenini-moore:2000:INLG}. 
Information retrieval systems are recently developed to extract arguments relevant to a given debate motion~\cite{sato-EtAl:2015:ACL-IJCNLP-2015-System-Demonstrations}. Although content ordering has been investigated~\cite{reisert-EtAl:2015:ARG-MINING,yanase-EtAl:2015:ARG-MINING}, the output arguments are usually a collection of sentences from heterogeneous information sources, thus lacking coherence and conciseness. 
Our work aims to close the gap by generating eloquent and coherent arguments, assisted by an argument retrieval system. 

Recent progress in sequence-to-sequence (seq2seq) text generation models has delivered both fluent and content rich outputs by explicitly conducting content selection and ordering~\cite{gehrmann-etal-2018-bottom,wiseman-etal-2018-learning}, which is a promising avenue for  enabling end-to-end counter-argument construction~\cite{W18-5215}. 
In particular, our prior work~\cite{P18-1021} leverages passages retrieved from Wikipedia to improve the quality of generated arguments, yet Wikipedia itself has the limitation of containing mostly facts. 
By leveraging Wikipedia and popular news media, our proposed pipeline can enrich the factual evidence with high-quality opinions and reasoning.

Our work is also in line with argument retrieval research, 
where prior effort mostly considers single-origin information source~\cite{rinott-EtAl:2015:EMNLP,C18-1176,W17-5106,P18-1023}. 
Recent work by \newcite{N18-5005} indexes all web documents collected in Common Crawl, which inevitably incorporates noisy, low-quality content. 
Besides, existing work treats individual sentences as arguments, 
disregarding their crucial discourse structures and logical relations with adjacent sentences.  
Instead, we use multiple high-quality information sources, 
and construct paragraph-level passages to retain the context of arguments.

\begin{figure*}[t]
\centering
  \includegraphics[width=155mm]{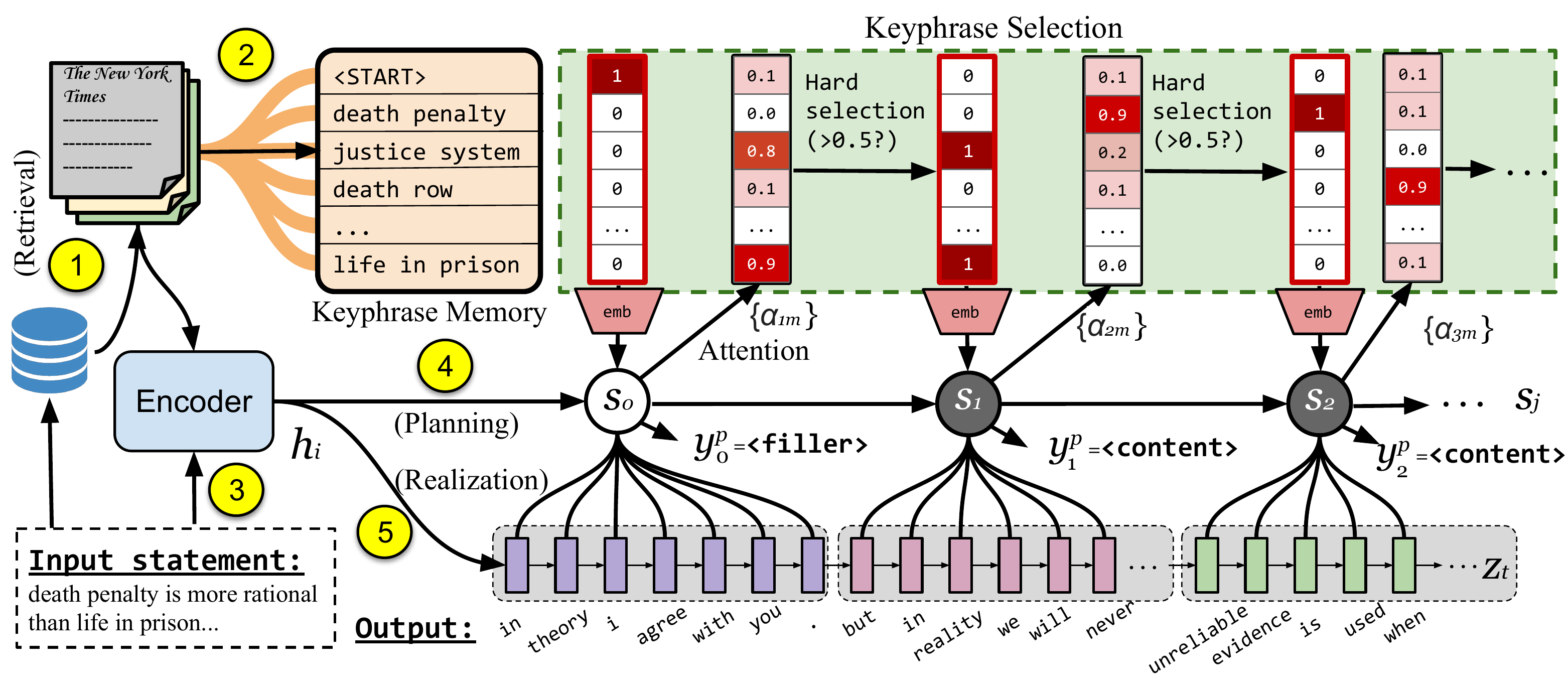}
  \caption{\fontsize{10}{12}\selectfont 
  Architecture of {\sc CANDELA}. 
  {\footnotesize \circled{1}} Argument retrieval (\S~\ref{sec:retrieval}): a set of passages are retrieved and ranked based on relevance and stance (\S~\ref{subsec:search},~\ref{subsec:rank}), from which {\footnotesize \circled{2}} a set of keyphrases are extracted (\S~\ref{subsec:kp}), with both as input for argument generation. 
  {\footnotesize \circled{3}} The biLSTM encoder consumes the input statement and passages returned from step 1. 
  {\footnotesize \circled{4}} A text planning decoder outputs a representation per sentence, and simultaneously predicts an argumentative function and selects keyphrases to include for the next sentence to be generated (\S~\ref{subsec:planner}). 
  {\footnotesize \circled{5}} A content realization decoder produces the counter-argument (\S~\ref{subsec:realizer}).
  }
  \vspace{-3mm}
  \label{fig:pipeline}
\end{figure*}

%% file: 030overview.tex
Our counter-argument generation framework, as shown in Figure~\ref{fig:pipeline}, has two main components: \textbf{argument retrieval} model (\S~\ref{sec:retrieval}) that takes the input statement and a search engine, and outputs relevant passages and keyphrases, which are used as input for our \textbf{argument generation} model (\S~\ref{sec:model}) to produce a fluent and informative argument. 

Concretely, the argument retrieval component retrieves a set of candidate passages from Wikipedia and news media (\S~\ref{subsec:search}), then further selects passages according to their stances towards the input statement (\S~\ref{subsec:rank}).
A \textbf{keyphrase extraction} module distills the refined passages into a set of talking points, which comprise the keyphrase memory as additional input for generation (\S~\ref{subsec:kp}). 

The argument generation component first runs the \textbf{text planning decoder} (\S~\ref{subsec:planner}) to produce a sequence of hidden states, each corresponding to a {\it sentence-level representation} that encodes the selection of keyphrases to cover, as well as the predicted argumentative function for a desired language style.
The \textbf{content realization decoder} (\S~\ref{subsec:realizer}) then generates the argument conditioned on the sentence representations.



%% file: 040retrieval.tex
\subsection{Information Sources and Indexing}
\label{subsec:search}

We aim to build a search engine from diverse information sources with factual evidence and varied opinions of high quality. 
To achieve that, we use Common Crawl\footnote{\url{http://commoncrawl.org/}} to collect a large-scale online news dataset covering four major English news media: \texttt{The New York Times} (NYT), \texttt{The Washington Post} (WaPo), \texttt{Reuters}, and \texttt{The Wall Street Journal} (WSJ). HTML files are processed using the open-source tool jusText~\cite{pomikalek2011removing} 
to extract article content. 
We deduplicate articles and remove the ones with less than $50$ words. 
We also download a Wikipedia dump. 
About $12$ million articles are processed in total, with basic statistics shown in Table \ref{tab:media}.

We segment articles into passages with a sliding window of three sentences, with a step size of two. We further constraint the passages to have at least $50$ words. For shorter passages, we keep adding subsequent sentences until reaching the length limit.
Per Table~\ref{tab:media}, $120$ million passages are preserved and indexed with Elasticsearch~\cite{gormley2015elasticsearch}
as done in~\newcite{N18-5005}. 


\begin{table}[t]
\fontsize{9}{11}\selectfont
 \setlength{\tabcolsep}{1.0mm}
  \centering
    \begin{tabular}{lccc}
        \toprule
        Source & \# Articles & \# Passages & Date Range \\
        \midrule
        \textbf{Wikipedia} & 5,743,901 & 42,797,543 & dump of 12/2016 \\
        \textbf{WaPo} & 1,109,672 & 22,564,532 & 01/1997 - 10/2018 \\
        \textbf{NYT} & 1,952,446 & 28,904,549 & 09/1895 - 09/2018 \\ 
        \textbf{Reuters} & 1,052,592 & 9,913,400 & 06/2005 - 09/2018 \\
        \textbf{WSJ} & 2,059,128 & 16,109,392 & 01/1996 - 09/2018 \\
        \hdashline
        Total & 11,917,739 & 120,289,416 & - \\
       \bottomrule
    \end{tabular}
     \caption{
    Statistics on information sources for argument retrieval. News media are sorted by ideological leanings from left to right, according to \url{https://www.adfontesmedia.com/}.
  \label{tab:media}}
\end{table}

\medskip
\noindent \textbf{Query Formulation.} 
For an input statement with multiple sentences, one query is constructed per sentence, if it has more than $5$ content words ($10$ for questions), and at least $3$ are distinct. 
For each query, the top $20$ passages ranked by BM25~\cite{robertson1995okapi} are retained, per medium.
All passages retrieved for the input statement are merged and deduplicated, and they will be ranked as discussed in \S~\ref{subsec:rank}.

\subsection{Keyphrase Extraction}
\label{subsec:kp}
Here we describe a keyphrase extraction procedure for both input statements and retrieved passages, which will be utilized for passage ranking as detailed in the next section. 

For {\it input statement}, our goal is to identify a set of phrases representing the issues under discussion, such as ``death penalty'' in Figure~\ref{fig:motivating_example}. 
We thus first extract the topic signature words~\cite{C00-1072} for input representation, and expand them into phrases that better capture semantic meanings. 

Concretely, topic signature words of an input statement are calculated against all input statements in our training set with log-likelihood ratio test. In order to cover phrases with related terms, we further expand this set with their synonyms, hyponyms, hypernyms, and antonyms based on WordNet~\cite{H94-1111}. 
The statements are first parsed with Stanford part-of-speech tagger~\cite{manning-EtAl:2014:P14-5}. Then regular expressions are applied to extract candidate noun phrases and verb phrases (details in Appendix~\ref{subsec:kpextraction}). 
A keyphrase is selected if it contains: (1) at least one content word, (2) no more than 10 tokens, and (3) at least one topic signature word or a Wikipedia article title.

For {\it retrieved passages}, their keyphrases are extracted using the same procedure as above, except that the input statement's topic signature words are used as references again.

\subsection{Passage Ranking and Filtering}
\label{subsec:rank}

We merge the retrieved passages from all media and rank them based on the number of words in overlapping keyphrases with the input statement. To break a tie, with the input as the reference, we further consider the number of its topic signature words that are covered by the passage, then the coverage of non-stopword bigrams and unigrams. In order to encourage diversity, we discard a passage if more than $50\%$ of its content words are already included by a higher ranked passage. 
In the final step, we filter out passages if they have the same stance as the input statement for given topics. We determine the stances of passages by adopting the {\bf stance scoring model} proposed by \newcite{E17-1024}. More details can be found in Appendix~\ref{subsec:stance}. 

%% file: 050model.tex
\subsection{Task Formulation} 
Given an input statement $\bm{X}=\{x_i\}$, a set of passages, and a keyphrase memory $\mathcal{M}$, our goal is to generate a counter-argument $\bm{Y}=\{y_t\}$ of a different stance as $\bm{X}$, $x_i$ and $y_t$ are tokens at timestamps $i$ and $t$. 
Built upon the sequence-to-sequence (seq2seq) framework with input attention~\cite{NIPS2014_5346,bahdanau2014neural}, the input statement and the passages selected in \S~\ref{sec:retrieval} are encoded by a bidirectional LSTM (biLSTM) encoder into a sequence of hidden states $\bm{h}_i$. 
The last hidden state of the encoder is used as the first hidden state of both text planning decoder and content realization decoder.

As depicted in Figure \ref{fig:pipeline}, the counter-argument is generated as follows. A text planning decoder (\S~\ref{subsec:planner}) first calculates a sequence of sentence representations $\bm{s}_j$ (for the $j$-{th} sentence) by encoding the keyphrases selected from the previous timestamp $j-1$. 
During this step, an {\it argumentative function} label is predicted to indicate a desired language style for each sentence, and a subset of the keyphrases are selected from $\mathcal{M}$ (\textit{content selection}) for the next sentence. 
In the second step, a content realization decoder (\S~\ref{subsec:realizer}) generates the final counter-argument conditioned on previously generated tokens and the corresponding sentence representation $\bm{s}_j$.

\subsection{Text Planning Decoder}
\label{subsec:planner}
Text planning is an important component for natural language generation systems to decide on content structure for the target generation~\cite{A97-1039,reiter2000building}. 
We propose a text planner with two objectives: selecting talking points from the keyphrase memory $\mathcal{M}$, and choosing a proper argumentative function per sentence. 
Concretely, we train a sentence-level LSTM that learns to generate a sequence of sentence representations $\{\bm{s}_j\}$ given the selected keyphrase set $\mathbb{C}(j)$ as input for the $j$-th sentence: 

{\fontsize{10}{11}\selectfont
\setlength{\abovedisplayskip}{2pt}
\setlength{\belowdisplayskip}{2pt}
\begin{align}
 	 & \bm{s}_j = f(\bm{s}_{j-1}, \sum_{e_k \in \mathbb{C}(j)} \bm{e}_k) \label{eq:sp-states} 
\end{align}
}
where $f$ is an LSTM network, $\bm{e}_k$ is the embedding for a selected phrase, represented by summing up all its words' Glove embeddings~\cite{pennington-socher-manning:2014:EMNLP2014} in our experiments. 

\medskip
\noindent \textbf{Content Selection $\mathbb{C}(j)$.} 
We propose an attention mechanism to conduct content selection and yield $\mathbb{C}(j)$ from the representation of the previous sentence $\bm{s}_{j-1}$ to encourage topical coherence. 
To allow the selection of multiple keyphrases, we use the sigmoid function to calculate the score: 

{\fontsize{10}{11}\selectfont
\setlength{\abovedisplayskip}{2pt}
\setlength{\belowdisplayskip}{2pt}
\begin{align}
 	 & \alpha_{jm} = \text{sigmoid}(\bm{e}_m \bm{W}^{pa} \bm{s}_{j-1})  \label{eq:attn_2} 
\end{align}
} 

where $\bm{W}^{pa}$ are trainable parameters, keyphrases with $\alpha_{jm} > 0.5$ are included in $\mathbb{C}(j)$, and the keyphrase with top attention value is always selected. 
We further prohibit a keyphrase from being chosen for more than once in multiple sentences. 
For the first sentence $\bm{s}_0$, $\mathbb{C}(0)$ only contains $\texttt{<start>}$, whose embedding is randomly initialized. 
During training, 
the {\it true labels} of $\mathbb{C}(j)$ are constructed as follows: a keyphrase in $\mathcal{M}$ is selected for the $j$-th gold-standard argument sentence if they overlap with any content word.

\medskip
\noindent \textbf{Argumentative Function Prediction $y^p_j$.} 
As shown in Figure~\ref{fig:motivating_example}, humans often deploy stylistic languages to achieve better persuasiveness, e.g. agreement as a concessive move. We aim to inform the realization decoder about the choice of style, and thus distinguish between two types of argumentative functions: \textbf{argumentative content sentence} which delivers the critical ideas, e.g. ``{\it unreliable evidence is used when there is no witness}'', and \textbf{argumentative filler sentence} which contains stylistic languages or general statements (e.g., ``{\it you can't bring dead people back to life}'').

Since we do not have argumentative function labels, during training, we use the following rules to automatically label each sentence as {\it content sentence} if it has at least $10$ words ($20$ for questions) and satisfy the following conditions: (1) it has at least two topic signature words of the input statement or a gold-standard counter-argument\footnote{When calculating topic signatures for gold-standard arguments, all replies in the training set are used as background.}, or (2) at least one topic signature word with a discourse marker at the beginning of the sentence. If the first three words in a {\it content sentence} contain a pronoun, the previous sentence is labeled as such too. Discourse markers are selected from PDTB discourse connectives (e.g., \textit{as a result}, \textit{eventually}, or \textit{in contrast}). The full list is included in Appendix~\ref{subsec:discourse}. 
All other sentences become {\it filler sentences}. 
In the future work, we will consider utilizing learning-based methods, e.g., \newcite{W17-5102}, to predict richer argumentative functions.

The argumentative function label $y_j^p$ for the $j$-th sentence is calculated as follows: 

{\fontsize{10}{11}\selectfont
\begin{align}
\begin{split}
 P(y^p_j& | y^p_{<j}, \bm{X}) =  \\
        & \text{softmax}(\bm{w}_p^T(\tanh{(\bm{W}^{po}[\bm{c}_j;\bm{s}_j])}) + \bm{b}_p) \label{eq:output_1} \\
\end{split}\\[-0.0em]
\begin{split}
 & \bm{c}_j = \sum_{e_m \in \mathcal{M}} \alpha_{jm} \bm{e}_m \label{eq:attn_1} 
\end{split}
\end{align}
} 
where $\alpha_{jm}$ is the alignment score computed as in Eq.~\ref{eq:attn_2}, $\bm{c}_j$ is the attention weighted context vector, $\bm{w}_p$, $\bm{W}^{po}$, and $\bm{b}_p$ are trainable parameters.

\subsection{Content Realization Decoder}
\label{subsec:realizer}
The content realization decoder generates the counter-argument word by word, with another LSTM network $f^w$. We denote the sentence id of the $t$-th word in the argument as $J(t)$, then the sentence representation $\bm{s}_{J(t)}$ from the text planning decoder, together with the embedding of the previous generated token $\bm{y}_{t-1}$, are fed as input to calculate the hidden state $\bm{z}_t$:

{\fontsize{10}{11}\selectfont
\begin{align}
 	 & \bm{z}_t = f^w(\bm{z}_{t-1}, \tanh(\bm{W}^{wp}\bm{s}_{J(t)} + \bm{W}^{ww}\bm{y}_{t-1} + \bm{b}^{w})) \label{eq:r-states}
\end{align}
}

The conditional probability of the next token $y_t$ is then computed over a standard softmax, with an attention mechanism applied on the encoder hidden states $\bm{h}_i$ to obtain the context vector $\bm{c}_t^w$:

{\fontsize{10}{11}\selectfont
\begin{align}
\begin{split}
 P(y_t&|y_{<t}, \bm{X}, \bm{s}_{J(t)}) = \\
        & \text{softmax}(\bm{w}_w^T(\tanh{(\bm{W}^{wo}[\bm{c}^w_t;\bm{z}_t])}) + \bm{b}^o) \label{eq:output_2} \\
\end{split}\\[-0.0em]
\begin{split}
 & \bm{c}^w_t = \sum_{i=1}^{|\bm{X}|} \beta_{ti} \bm{h}_i 
\end{split}\\[-0.0em]
\begin{split}
 & \beta_{ti} = \text{softmax}(\bm{h}_i \bm{W}^{wa} \bm{z}_t)  
\end{split}
\end{align}
} 
where $\beta_{ti}$ is the input attention, 
$\bm{W}^{wp}$, $\bm{W}^{ww}$, $\bm{W}^{wo}$, $\bm{W}^{wa}$, $\bm{b}^o$, $\bm{w}_w$, and $\bm{b}^w$ are learnable.

\medskip
\noindent \textbf{Reranking-based Beam Search.} 
Our content realization decoder utilizes beam search enhanced with a reranking mechanism, where we sort the beams at the end of each sentence by the number of selected keyphrases that are generated. We also discard beams with $n$-gram repetition for $n\geq 4$.

\subsection{Training Objective}
Given all model parameters $\theta$, our mixed objective considers the target argument ($\mathcal{L}_{\text{arg}}(\theta)$), the argumentative function type ($\mathcal{L}_{\text{func}}(\theta)$), and the next sentence keyphrase selection ($\mathcal{L}_{\text{sel}}(\theta)$):

{\fontsize{9.5}{11}\selectfont
\begin{align}
\begin{split}
  & \mathcal{L}(\theta) = \mathcal{L}_{\text{arg}}(\theta) + \gamma \cdot \mathcal{L}_{\text{func}}(\theta) + \eta \cdot \mathcal{L}_{\text{sel}}(\theta) \\
\end{split}\\[-0.0em]
\begin{split}
  & \mathcal{L}_{\text{arg}}(\theta) = -{\sum_{(\bm{X},\bm{Y})\in D}} \log P(\bm{Y}|\bm{X};\theta) \\ 
\end{split}\\[-0.5em]
\begin{split}
  & \mathcal{L}_{\text{func}}(\theta)= -{\sum_{(\bm{X},\bm{Y}^p)}} \log P(\bm{Y}^{p}|\bm{X};\theta) \\
\end{split}\\[-0.5em]
\begin{split}
  & \mathcal{L}_{\text{sel}}(\theta)= \\
  & -\sum_{\bm{Y}^p}\sum_{j=1}^{|\bm{Y}^p|}(\sum_{e_m \in \mathbb{C}(j)} \log(\alpha_{jm}) + \sum_{e_m \not\in \mathbb{C}(j)} \log (1 - \alpha_{jm}))\label{eq:loss_overall} \\
\end{split}
\end{align}
} 
where $D$ is the training corpus,  $(\bm{X}, \bm{Y})$ are input statement and counter-argument pairs, and $\bm{Y}^p$ are the sentence function labels. $\alpha_{jm}$ are keyphrase selection labels as computed in Eq.~\ref{eq:attn_2}. 
For simplicity, we set $\gamma$ and $\eta$ as $1.0$ in our experiments, while they can be further tuned as hyper-parameters.

%% file: 060experiment.tex


\subsection{Data Collection and Preprocessing}

We use the same methodology as in our prior work~\cite{P18-1021} to collect an argument generation dataset from Reddit \texttt{/r/ChangeMyView}.\footnote{We further crawled $42,649$ threads from July 2017 to December 2018, compared to the previously collected dataset.}  
To construct input statement and counter-argument pairs, we treat the original poster (OP) of each thread as the input. We then consider the high quality root replies, defined as the ones awarded with $\Delta$s or with more upvotes than downvotes (i.e., \texttt{karma} $> 0$). 
It is observed that each paragraph often makes a coherent argument. 
Therefore, these replies are broken down into paragraphs, 
and a paragraph is retained as a target argument to the OP if it has more than $10$ words and at least one argumentative content sentence.

We then identify threads in the domains of politics and policy, and remove posts with offensive languages.
Most recent threads are used as test set. As a result, we have $11,356$ threads or OPs ($217,057$ arguments) for training, $1,774$ ($33,318$ arguments) for validation, and $1,703$ ($36,777$ arguments) for test. They are split into sentences and then tokenized by the Stanford CoreNLP toolkit~\cite{manning-EtAl:2014:P14-5}.

\smallskip
\noindent \textbf{Training Data Construction for Passages and Keyphrase Memory.} 
Since no gold-standard annotation is available for the input passages and keyphrases, we acquire training labels by constructing queries from the gold-standard arguments as described in \S~\ref{subsec:search}, and reranking retrieved passages based on the following criteria in order: 
(1) coverage of topic signature words in the input statement; 
(2) a weighted summation of the coverage of $n$-grams in the argument\footnote{We choose $0.5, 0.3, 0.2$ as weights for $4$-grams, trigrams, and bigrams, respectively.}; 
(3) the magnitude of stance score, where we keep the passages of the same polarity as the argument; 
(4) content word overlap with the argument; 
and (5) coverage of topic signature words in the argument.



\subsection{System and Oracle Retrieved Passages}
For evaluation, we employ both {\it system} retrieved passages (i.e.,  constructing queries from OP) and KM (\S~\ref{sec:retrieval}), and {\it oracle} retrieved passages (i.e., constructing queries from target argument) and KM as described in training data construction. 
Statistics on the final dataset are listed in Table \ref{tab:stats}. 

\begin{table}[ht]
\fontsize{10}{11}\selectfont
 \setlength{\tabcolsep}{1.0mm}
  \centering
    \begin{tabular}{llll}
        \toprule
        & {\it Training} & \textit{System} & \textit{Oracle} \\
        \midrule
        Avg. \# words per OP & 383.7 & 373.0 & 373.0 \\
        Avg. \# words per argument  & 66.0 & 65.1 & 65.1  \\
        Avg. \# passage & 4.3 & 9.6 & 4.2 \\
        Avg. \# keyphrase  & 57.1& 128.6 & 56.6\\
        \bottomrule
    \end{tabular}
    \caption{
    Statistics on the datasets for experiments. 
    }
    \label{tab:stats}
    
\end{table}

\subsection{Comparisons}
\label{subsec:comparison}
In addition to a \textsc{\bf Retrieval} model, where the top ranked passage is used as counter-argument, 
we further consider four systems for comparison.  
(1) A standard \textsc{\bf Seq2seq} model with attention, where we feed the OP as input and train the model to generate counter-arguments. 
Regular beam search with the same beam size as our model is used for decoding. 
(2) A \textsc{\bf Seq2seqAug} model with additional input of the keyphrase memory and ranked passages, both concatenated with OP to serve as the encoder input. The reranking-based decoder in our model is also implemented for \textsc{Seq2seqAug} to enhance the coverage of input keyphrases. 
(3) An ablated \textsc{Seq2seqAug} model where the passages are removed from the input. 
(4) We also reimplement the argument generation model in our prior work~\cite{P18-1021} ({\bf H\&W}) with PyTorch~\cite{paszke2017pytorch},  which is used for \textsc{CANDELA} implementation.  H$\&$W takes as input the OP and ranked passages, and then uses two separate decoders to first generate all keyphrases and then the counter-argument. 
%
For our model, we also implement a variant where the input only contains the OP and the keyphrase memory.

\subsection{Training Details}
For all models, we use a two-layer LSTM for all encoders and decoders with a dropout probability of $0.2$ between layers~\cite{NIPS2016_6241}. All layers have $512$-dimensional hidden states.
We limit the input statement to $500$ tokens, the ranked passages to $400$ tokens, and the target counter-argument to $120$ tokens. 
Our vocabulary has $50K$ words for both input and output, with $300$-dimensional word embeddings initialized with GloVe~\cite{pennington-socher-manning:2014:EMNLP2014} and fine-tuned during model training. 
We use AdaGrad~\cite{duchi2011adaptive} with a learning rate of $0.15$ and an initial accumulator of $0.1$ as the optimizer, with the gradient norm clipped to $2.0$. Early stopping is implemented according to the perplexity on validation set. 
For all our models the training takes approximately $30$ hours ($40$ epochs) on a Quadro P5000 GPU card, with a batch size of $64$. 
For beam search, we use a beam size of $5$, tuned from $\{5, 10, 15\}$ on validation.

We also pre-train a biLSTM for encoder based on all OPs from the training set, and an LSTM for content realization decoder based on two sources of data: 
$353$K counter-arguments that are high quality root reply paragraphs extended with posts of non-negative karma, and $2.4$ million retrieved passages randomly sampled from the training set. Both are trained as done in~\newcite{bengio2003neural}. We then use the first layer's parameters to initialize all models, including our comparisons. 

%% file: 070analysis.tex



\subsection{Automatic Evaluation} 

\begin{table*}[t]
\centering
\fontsize{10}{11}\selectfont
 \setlength{\tabcolsep}{1.0mm}
\centering
    \begin{tabular}{l llllll l llllll}
    \toprule
        & \multicolumn{6}{c}{\textit{w/ System Retrieval}} & \phantom{} & \multicolumn{6}{c}{\textit{w/ Oracle Retrieval}} \\
        \cmidrule{2-7} \cmidrule{9-14}
        & \textbf{B-2} & \textbf{B-4} & \textbf{R-2} & \textbf{MTR} & \textbf{\#Word} & \textbf{\#Sent} &
        \phantom{}  & \textbf{B-2} & \textbf{B-4}& \textbf{R-2} & \textbf{MTR}  & \textbf{\#Word} & \textbf{\#Sent} \\
        \midrule
        \textsc{Human} & - & - & - & - & 66 & 22 &\phantom{} & - & - & - & - & 66 & 22 \\
         \textsc{Retrieval}  & 7.55 & 1.11 & 8.64 & 14.38 &  123 & 23   &\phantom{} & 10.97 & 3.05 & 23.49 & 20.08  & 140 & 21 \\
         \midrule
        \multicolumn{12}{l}{\bf Comparisons} \\
        \textsc{Seq2seq} & 6.92 & 2.13 & 13.02 & 15.08 & 68 & 15 &\phantom{}&  6.92 & 2.13 & 13.02 & 15.08 & 68 & 15  \\
        \textsc{Seq2seqAug} & 8.26 & 2.24 & 13.79 & 15.75 & 78 & 14 &\phantom{}& 10.98 & 4.41 & 22.97 & 19.62 & 71 & 14 \\
        \quad\quad \textit{w/o psg} & 7.94 & 2.28 & 10.13 & 15.71 & 75 & 12 &\phantom{}& 9.89 & 3.34 & 14.20  & 18.40 & 66 & 12 \\
        H\&W~\shortcite{P18-1021} & 3.64 & 0.92 & 8.83 & 11.78 & 51 & 12 &\phantom{} & 8.51  & 2.86 & 18.89 & 17.18 & 58 & 12 \\
        \midrule
        \multicolumn{12}{l}{\bf Our Models} \\
       \textsc{CANDELA}   & 12.02$^\ast$ & {\bf 2.99}$^\ast$ & {\bf 14.93}$^\ast$ & {\bf 16.92}$^\ast$ & 119 & 22 &\phantom{}& 15.80$^\ast$ & {\bf 5.00}$^\ast$ & {\bf 23.75} & {\bf 20.18} & 116 & 22 \\
       \quad\quad  \textit{w/o psg}  & {\bf 12.33}$^\ast$ & 2.86$^\ast$ & 14.53$^\ast$ & 16.60$^\ast$ & 123& 23 &\phantom{}& {\bf 16.33}$^\ast$ & 4.98$^\ast$ & 23.65 & 19.94 & 123 & 23\\
     \bottomrule
    \end{tabular}
    \caption{
    Main results on argument generation. We report BLEU-2 (B-2), BLEU-4 (B-4), ROUGE-2 (R-2) recall, METEOR (MTR), and average number of words per argument and per sentence. Best scores are in bold. $\ast$: statistically significantly better than all comparisons (randomization approximation test~\cite{noreen1989computer}, $p<0.0005$). Input is the same for \textsc{Seq2seq} for both system and oracle setups.
  }
  \label{tab:main-results}
\end{table*}

We employ ROUGE~\cite{W04-1013}, a recall-oriented metric, BLEU~\cite{papineni-EtAl:2002:ACL}, based on $n$-gram precision, and METEOR~\cite{denkowski-lavie:2014:W14-33}, measuring unigram precision and recall by considering synonyms, paraphrases, and stemming. 
BLEU-2, BLEU-4, ROUGE-2 recall, and METEOR are reported in Table \ref{tab:main-results} for both setups. 

Under system setup, our model \textsc{CANDELA} statistically significantly outperforms all comparisons and the retrieval model in all metrics, based on a randomization test~\cite{noreen1989computer} ($p < 0.0005$). Furthermore, our model generates longer sentences whose lengths are comparable with human arguments, both with about $22$ words per sentence. This also results in longer arguments. 
Under oracle setup, all models are notably improved due to the higher quality of reranked passages, and our model achieves statistically significantly better BLEU scores. 
Interestingly, we observe a decrease of ROUGE and METEOR, but a marginal increase of BLEU-2 by removing passages from our model input. 
This could be because the passages introduce divergent content, albeit probably on-topic, that cannot be captured by BLEU. 

\begin{figure}[t]
\centering
\includegraphics[width=70mm]{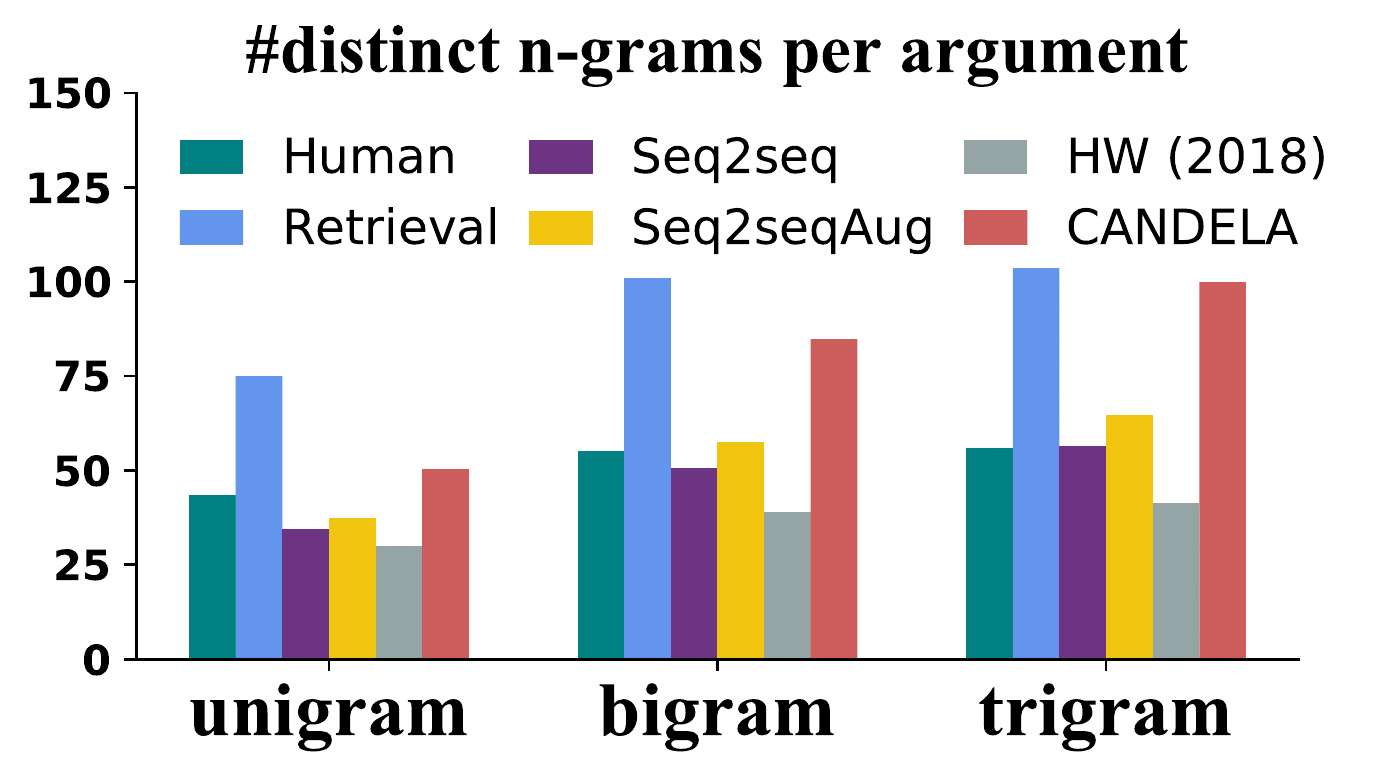}
\caption{
Average number of distinct $n$-grams per argument.
}
\label{fig:dist-ngram}
\end{figure}

\begin{figure}[t]
\fontsize{10}{12}\selectfont
    \centering
    \begin{tabular}{|l|cccc|}
    \hline
    & \multicolumn{4}{c|}{\it K} \\
    & 100 & 500 & 1000 & 2000 \\
    \textsc{Human} & \cellcolor{blue!44}44.1 & \cellcolor{blue!26}25.8 & \cellcolor{blue!19}18.5 & \cellcolor{blue!12}12.0 \\
    \textsc{Retrieval} & \cellcolor{blue!51}50.6 & \cellcolor{blue!33}33.3 & \cellcolor{blue!26}26.0 & \cellcolor{blue!19}18.6 \\
    \hline
    \textsc{Seq2seq} & \cellcolor{blue!26}25.0 & \cellcolor{blue!8}7.5 & \cellcolor{blue!3}3.2 & \cellcolor{blue!1}1.2 \\
    \textsc{Seq2seqAug}& \cellcolor{blue!28}28.2 & \cellcolor{blue!9}9.2 & \cellcolor{blue!5}4.6 & \cellcolor{blue!2}1.8 \\
    H\&W~\shortcite{P18-1021} & \cellcolor{blue!39}38.6 & \cellcolor{blue!24}24.0 & \cellcolor{blue!20}19.5 & \cellcolor{blue!16}16.2 \\
    \textsc{CANDELA}  & \cellcolor{blue!30}30.0 & \cellcolor{blue!11}10.5 & \cellcolor{blue!5}5.3 & \cellcolor{blue!2}2.3 \\
   \hline
    \end{tabular}
    \caption{Percentage of words in arguments that are not in the top-$K$ ($K = 100, 500, 1000, 2000$) frequent words seen in training. Darker color indicates higher portion of uncommon words found in the arguments.
    }
    \label{fig:topk}
\end{figure}

\medskip
\noindent \textbf{Content Diversity.} 
We further measure whether our model is able to generate diverse content. 
First, borrowing the diversity measurement from dialogue generation research~\cite{li-EtAl:2016:N16-11}, we report the average number of distinct $n$-grams per argument under system setup in Figure~\ref{fig:dist-ngram}. Our system generates more unique unigrams and bigrams than other automatic systems, 
underscoring its capability of generating diverse content. Our model also maintains a comparable type-token ratio (TTR) compared to systems that generate shorter arguments, e.g., a $0.79$ for bigram TTR of our model versus $0.83$ and $0.84$ for \textsc{Seq2seqAug} and \textsc{Seq2seq}. 
\textsc{Retrieval}, containing top ranked passages of human-edited content, produces the most distinct words. 



Next, we compare how each system generates content beyond the common words. 
As shown in Figure~\ref{fig:topk}, human-edited text, including gold-standard arguments (\textsc{Human}) and retrieved passages, tends to have higher usage of uncommon words than automatic systems, suggesting the gap between human vs. system arguments. 
Among the four automatic systems, our prior model~\cite{P18-1021} generates a significantly higher portion of uncommon words, yet further inspection shows that the output often includes more off-topic information.

\subsection{Human Evaluation}
Human judges are asked to rate arguments on a Likert scale of 1 (worst) to 5 (best) on the following three aspects: 
\textbf{grammaticality}---denotes language fluency; 
\textbf{appropriateness}---indicates if the output is on-topic and on the opposing stance; 
\textbf{content richness}---measures the amount of distinct talking points. 
In order to promote consistency of annotation, we provide descriptions and sample arguments for each scale. For example, an appropriateness score of $3$ means the counter-argument contains  relevant words and is likely to be on a different stance. 
The judges are then asked to rank all arguments for the same input based on their overall quality.

We randomly sampled $43$ threads from the test set, and hired three native or proficient English speakers to evaluate arguments generated by \textsc{Seq2seqAug}, our prior argument generation model (H$\&$W), and the new model \textsc{CANDELA}, along with gold-standard \textsc{Human} arguments and the top passage by \textsc{Retrieval}. 

\begin{table}[t]
\fontsize{9.5}{11}\selectfont
 \setlength{\tabcolsep}{1.0mm}
  \centering
    \begin{tabular}{lcc lcc}
        \toprule
        & {\bf Gram.} & {\bf Appr.} & {\bf Cont.} & {\bf Top-1} & {\bf Top-2} \\
        \midrule
        \textsc{Human} & 4.95  & 4.23 & 4.39 & 75.8\% & 85.8\%  \\
        \textsc{Retrieval} & 4.85  & 3.04  & 3.68  & 17.5\% & 55.8\%\\
        \midrule
        \textsc{Seq2seqAug} & {\bf 4.83}& 2.67 & 2.47 & 1.7\% & 22.5\%\\
         H\&W \shortcite{P18-1021} & 3.86& 2.27 & 2.10  & 1.7\% & 7.5\% \\
         \textsc{CANDELA}  & 4.59 & {\bf 2.97} & {\bf 2.93$^\ast$} & 3.3\% & 28.3\%\\
        \bottomrule
        
    \end{tabular}
    \caption{\fontsize{10}{12}\selectfont  
    Human evaluation on grammaticality (Gram), appropriateness (Appr), and content richness (Cont.), on a scale of 1 to 5 (best). 
    The best result among automatic systems is highlighted in bold, with statistical significance marked with $\ast$ (approximation randomization test, $p<0.0005$). 
    The highest standard deviation among all is $1.0$. 
    Top-1/2: $\%$ of evaluations a system being ranked in top 1 or 2 for overall quality.
  }
  \label{tab:human-eval}
\end{table}

\smallskip
\noindent \textbf{Results.} 
The first $3$ examples are used only for calibration, and the remaining $40$ are used to report results in Table~\ref{tab:human-eval}. 
Inter-annotator agreement scores (Krippendorff's $\alpha$) of $0.44$, $0.58$, $0.49$ are achieved for the three aspects, implying general consensus to intermediate agreement. 

Our system obtains the highest appropriateness and content richness among all automatic systems. 
This confirms the previous observation that our model produces more informative argument than other neural models. 
\textsc{Seq2seqAug} has a marginally better grammaticality score, likely due to the fact that our arguments are longer, and tend to contain less fluent generation towards the end. 

Furthermore, we see that human arguments are ranked as the best in about $76\%$ of the evaluation, followed by \textsc{Retrieval}. Our model is more likely to be ranked top than any other automatic models. Especially, our model is rated better than either \textsc{Human} or \textsc{Retrieval}, i.e., human-edited text, in $39.2\%$ of the evaluations, compared to $34.2\%$ for \textsc{Seq2seqAug} and $13.3\%$ for our prior model.

\subsection{Sample Arguments and Discussions}
We show sample outputs of different systems alongside human constructed counter-argument in Figure~\ref{fig:sample-outputs}. 
As can be seen, our system arguments cover many relevant talking points, including the idea of ``\textit{taking care of children}'' that is also used by human. 
It further illustrates the effectiveness of our retrieval system and the usage of keyphrase selection as part of text planning to guide argument generation. 
Moreover, we also observe that our model generation is able to switch between argumentative content sentence and filler sentence, though better control is needed to improve coherence. 
Meanwhile, \textsc{Seq2seq} frequently echos words from OP, and both \textsc{Seq2seq} and \textsc{Seq2seqAug} suffer from the problems of ``hallucination'' (e.g., the first sentence in \textsc{Seq2seqAug}) and repetition (e.g., the second and third sentences in \textsc{Seq2seq}).

\begin{figure}[t]
	\fontsize{9}{11}\selectfont
     \hspace{-2mm}
	\setlength{\tabcolsep}{0.8mm}
	\begin{tabular}{|p{75mm}|}
	\hline
	\textbf{OP}: 
    The wage gap isn't about gender. (...) 
    So one factor definitely the fact that there are some government jobs that are seriously dominated by women and happen to pay really low (...)\\

    
    \hline \hline 
    \textbf{Human}: Children are the first factor. I work in a traditionally male-dominated field, and questions are asked to determine familial status, even though those are illegal. Employers know single moms have to stay home if their child is sick. They know single moms can only work during daycare hours. They know single moms are unavailable for weekend and late night work. They know single moms cannot travel. The workplace is very family-unfriendly. 
    \\
    \hline
     \textbf{Seq2seq}: the problem with this is that there are a lot of people who do n't want to hire women . there are plenty of women who do n't have the money to pay for it . the problem is that women do n't need to pay for their work . they do n't have to worry about it . it 's a matter of money . \\
	
	\hline
	\textbf{Seq2seqAug}: men and women are not 39.5 \% of the pay gap . the problem is not that women are more likely to be victims of sexism , but rather that they are more natural good-looking/attractive action . this is not the case .  \\
	
    \hline
    \textbf{CANDELA}: the problem with this argument is that the {\color{red!70} \textbf{wage gap}} does not have to do with the {\color{blue!75} \textbf{gender pay gap}} . it is a fact that women are more likely to be able to {\color{green!70!blue} \textbf{take care  of their children}} than their {\color{pink!70!blue} \textbf{male counterparts}} . this is not to say that it is unfair to assume that women are being paid less than men , but that does not mean that it is not the case that {\color{orange!55} \textbf{women are discriminated against}} . {\it it is not a matter of the {{wage gap}} , it is a matter of opinion .} it is the job of the employer to make sure that the job is not the same as the other \\
    
    \begin{tabular}{p{3mm}p{70mm}}
     &
     \textbf{\underline{Keyphrase Memory}}: {\color{red!70} \textbf{wage gap}}; discrimination; {\color{blue!75} \textbf{gender pay  gaps}}; {\color{green!70!blue} \textbf{raise the child}}; {\color{pink!70!blue} \textbf{male colleagues}}; paid maternity leave;
    {\color{orange!55} \textbf{underlying gender discrimination}}
    \ldots
    \end{tabular}
    \\
    \hline

	\end{tabular}
	\caption{
   Sample arguments generated by different systems along with a sample human argument. For our model \textsc{CANDELA}, additionally shown are the keyphrase memory with selected phrases in color, and argumentative filler sentence in italics. 
   }
\label{fig:sample-outputs}
\end{figure}

Nonetheless, there is a huge space for  improvement. 
First, our model tends to overuse negation, such as \textit{``this is not to say that it is unfair...''}. It is likely due to its overfitting on specific stylistic languages, e.g., negation is often observed for refutation in debates~\cite{Q17-1016}. 
Second, human arguments have significantly better organization and often deploy complicated argumentation strategies~\cite{C18-1318}, which so far is not well captured by any automatic system. 
Both points inspire future work on (1) controlling of the language styles and corresponding content, and (2) mining argumentation structures for use in guiding generation with better planning.

%% file: 080conclusion.tex
We present a novel counter-argument generation framework, \textsc{CANDELA}. Given an input statement, it first retrieves arguments of different perspectives from millions of high-quality articles collected from diverse sources. An argument generation component then employs a text planning decoder to conduct content selection and specify a suitable language style at sentence-level, followed by a content realization decoder to produce the final argument. Automatic evaluation and human evaluation indicate that our model generates more proper arguments with richer content than non-trivial comparisons, with comparable fluency to human-edited content. 

%% file: 090appendix.tex
\subsection{Chunking Grammar for Keyhrase Extraction}
\label{subsec:kpextraction}

In order to construct keyphrase candidates, we compile a set of regular expressions based on the following grammar rules, and extract all matched \texttt{NP} and \texttt{VP} patterns as candidates.

{\fontsize{9}{11}
\centering
\vspace{3mm}
\begin{tabular}{|l|}
\hline
\texttt{NP: \{<DT|PP\$>?<JJ|JJR>*<NN.*|CD|JJ>+\}} \\
\texttt{PP: \{<IN><NP>\}} \\
\texttt{VP: \{<MD>?<VB.*><NP|PP>\}} \\
\hline
\end{tabular}}

\subsection{Stance Scoring Model}
\label{subsec:stance}

Our stance scoring model calculates the score by aggregating the sentiment words surrounding the opinion targets. Here we choose the keyphrases of input statement as opinion targets, denoted as $\mathbb{T}$. 
We then tally sentiment words, collected from~\newcite{hu2004mining}, towards targets in $\mathbb{T}$, with positive words counted as $+1$ and negative words as $-1$. Each score is discounted by $d_{\tau,l}^{-5}$, with $d_{\tau,l}$ being the distance between the sentiment word $l$ and the target $\tau \in \mathbb{T}$. The stance score of a text $psg$ (an input statement or a retrieved passage) towards opinion targets $\mathbb{T}$ is calculated as:

{
\begin{align}
        Q(psg, \mathbb{T}) = \sum_{\tau\in \mathbb{T}}\sum_{l\in psg} \text{sgn}(l) \cdot d^{-5}_{\tau,l} 
 \end{align}
}

In our experiments, we only keep passages with a stance score of the opposite sign to that of the input statement, and with a magnitude greater than $5$, i.e. $\vert Q(psg, \mathbb{T}) \vert > 5$ (determined by manual inspection on training set). 

\subsection{List of Discourse Markers}
\label{subsec:discourse}

As described in \S 5.2 in the main paper, we use a list of discourse markers together with topic signature words to label argumentative content sentences. The following list of discourse markers are manually selected from the Appendix B in \newcite{prasad2008penn}.
\begin{itemize}
    \item {\bf Contrast:} although, though, even though, by comparison, by contrast, in contrast, however, nevertheless, nonetheless, on the contrary, regardless, whereas
    \item {\bf Restatement/Equivalence/Generalization:} eventually, in short, in sum, on the whole, overall
    \item {\bf Result:} accordingly, as a result, as it turns out, consequently, finally, furthermore, hence, in fact, in other words, in short, in the end, in turn, therefore, thus, ultimately
\end{itemize}

\subsection{Human Evaluation Guideline}

Each human annotator is presented with 43 short argumentative text statements, where the first 3 statements are used as calibration for the annotator himself and excluded in the final study. 
The annotators are asked to evaluate 5 counter-arguments for every statement. 
For each counter-argument, they rate on a scale of 1 to 5 for the following aspects, and also specify the ranking among the 5 counter-arguments based on the overall quality. We display sample statements and score-level explanations in Table \ref{tab:human_eval}.

Three aspects of arguments to be evaluated:

\begin{itemize}
    \item {\bf Grammaticality:} whether the counterargument is fluent and has no grammar errors.
    \item {\bf Appropriateness:} whether the counterargument is on topic and on the right stance.
    \item {\bf Content Richness:} how many distinct talking points the counterargument conveys.
\end{itemize}

\begin{table*}
	\fontsize{10}{11}\selectfont
    \centering
    \begin{tabular}{lp{120mm}}
         \toprule
         \multicolumn{2}{p{120mm}}{\textbf{Statement:} Legislative bodies should be required to explain, in formal writing, why they voted a certain way when it comes to legislation. } \\
         \midrule
         \multicolumn{2}{c}{\textbf{Grammaticality:}} \\
         \rowcolor{lightgray!30}
          1 & With plenty of grammatical errors and not readable at all \\
          & e.g., \textit{``the way the way etc. 'm not 's important''} \\
          \rowcolor{lightgray!30}
          3 & With a noticeable amount of grammatical errors but is generally readable \\
          & e.g., \textit{``is a good example. i don't think should be the case. i're not going to talk whether or not it's a bad thing.''} \\
          \rowcolor{lightgray!30}
          5 &  With no grammatical errors at all and is clear to read \\
          & e.g., \textit{``i agree that the problem lies in the fact that too many representatives do n't understand the issues or have money influencing their decisions.''} \\
          \midrule
         \multicolumn{2}{c}{\textbf{Appropriateness:}} \\
         \midrule
         \rowcolor{lightgray!30}
         1 & Not relevant to the prompt at all \\
         & e.g., \textit{`` i don't think it 's fair to say that people should n't be able to care for their children''} \\
         \rowcolor{lightgray!30}
         2 & Remotely relevant to the prompt or relevant but poses an unclear stance, or contains obvious contradictions \\
         & e.g., \textit{``the problem with the current system is that there are many people who don't want to vote and they also don't want to vote.''} \\
         \rowcolor{lightgray!30}
         3 & Relevant to the prompt but stance is unclear \\
         & e.g., \textit{``i don't agree with you and i think legislative bodies do need to explain why they vote that way''} \\
         \rowcolor{lightgray!30}
         4 & Relevant to the prompt and is overall on the opposing stance with minor logical contradictions \\
         & e.g., \textit{``while i agree with you but i don't think it's a good idea for house reps to explain it because they have other work to do.''} \\
         \rowcolor{lightgray!30}
         5 & Relevant to the prompt and is on the opposing stance, has no unnatural repititions and logical contradictions \\
         & e.g., \textit{``there are hundreds of votes a year . how do you decide which ones are worth explaining ? so many votes are bipartisan if not nearly unanimous . do those all need explanations ? they only have two years right now and i do n't want them spending less time legislating .''} \\
          \midrule
         \multicolumn{2}{c}{\textbf{Content Richness:}} \\
         \midrule
         \rowcolor{lightgray!30}
         1 & Generic response with no useful information about the topic \\
         & e.g., \textit{``i do n't agree with your point about legislation but i 'm not going to change your view.''} \\
          \rowcolor{lightgray!30}
         3 & With one of two key information that are useful as counterargument \\
         & e.g., \textit{``i agree that this is a problem for congress term because currently it is too short.''} \\
         \rowcolor{lightgray!30}
         5 & With sufficient key information that are useful as counterargument \\
         & e.g., \textit{``congressional terms are too short and us hourse reps have to spend half of their time compaigning and securing campaign funds. they really have like a year worth of time to do policy and another year to meet with donors and do favors.''} \\
         \bottomrule
    \end{tabular}
    \caption{Sample statement with explanations on aspect scales. Due to the likely ambiguity in {\bf Appropriateness}, we provide explanations on every possible score. Example counter-arguments are also given alongside explanations.}
    \label{tab:human_eval}
\end{table*}

\subsection{Sample Output}

In Figure \ref{fig:sample-keyphrase} we show two sample snippets of our model outputs where reused keyphrases are highlighted in colors. 
Notice that even though in test time we disallow the same keyphrase to be selected more than once as the input for text planning decoder, the content realization decoder can still generate the same keyphrase multiple times. 
In this example, the phrase \textit{death penalty} is used three times across two sentences, while it is generally nature and relevant in the context.
We further include three complete sample outputs for different systems alongside the reranked passages from Figure \ref{fig:sample-reranked-1} to Figure \ref{fig:sample-reranked-3}.

\begin{figure*}[th]
    \def\arraystretch{1.5}
    \bgroup
    \def\arraystretch{1.5}
	\fontsize{9}{11}\selectfont
	\setlength{\tabcolsep}{0.8mm}
	\begin{tabular}{|p{160mm}|}
	\hline
	\textbf{Input}: 
    Gun control will do nothing to solve gun violence \\
    \hline
    \textbf{Keyphrase Memory}: {\color{red!70} \textbf{gun control}}; gun crime; {\color{blue!75} \textbf{reduce gun violence}}; homicide rate; {\color{green!75!blue} \textbf{gun violence}}; gun owners; american gun deaths; \ldots\\ 
    \hline
	\textbf{Model Output}:  
    \textit{I'm not sure if I'm going to try to change your view}, but I don't think it's accurate to say that {\color{red!70} \textbf{gun control}} is the only way to {\color{blue!75} \textbf{reduce gun violence}}. However, I think it's important to note that there is a significant amount of {\color{green!75!blue} \textbf{gun violence}} in the United States, which is why it's been used as a way to reduce crime \ldots \\
	\hline
	\hline
	\textbf{Input}: 
    The US should abolish the death penalty. It is morally wrong, economically inefficient, (\ldots) \\
    \hline
    \textbf{Keyphrase Memory}: {\color{red!70} \textbf{death penalty}}; capital punishment; {\color{blue!75} \textbf{killing someone}}; crime; {\color{green!75!blue} \textbf{murder}}; \ldots\\
    \hline
	\textbf{Model Output}:  
    the {\color{red!70} \textbf{death penalty}} is not the only thing that can be done in the first place . {\color{red!70} \textbf{death penalty}} is n't just about {\color{blue!75} \textbf{killing people}} , it 's about the fact that the {\color{red!70} \textbf{death penalty}} itself is not the same thing as {\color{green!75!blue} \textbf{murder}} . (\ldots) \\
	\hline
	\end{tabular}
\caption{\fontsize{10}{12}\selectfont 
Sample arguments by our model. 
We highlight the reused keyphrases from keyphrase memory with colors, and filler sentence with {\it italics}.} \label{fig:sample-keyphrase}
\egroup
\end{figure*}

%% file: 100samples.tex
\begin{figure*}[th]
    \bgroup
    \def\arraystretch{1.5}
	\fontsize{10}{11}\selectfont
    \hspace{-2mm}
    \centering
	\setlength{\tabcolsep}{0.8mm}
	\begin{tabular}{p{140mm}}

	\textbf{Input}: 
    The wage gap isn't about gender. (...) 
    So one factor definitely the fact that there are some government jobs that are seriously dominated by women and happen to pay really low (...)\\
    \hline
    \cellcolor{green!30}
    \textbf{Passage 1\quad Source: Wikipedia\quad Stance: -24.65 } \\
    \cellcolor{green!10}
    Research has found that women steer away from STEM fields because they believe they are not qualified for them; the study suggested that this could be fixed by encouraging girls to participate in more mathematics classes. One of the factors behind girls' lack of confidence might be unqualified or ineffective teachers. Teachers' gendered perceptions on their students' capabilities can create an unbalanced learning environment and deter girls from pursuing further STEM education. They can also pass these stereotyped beliefs unto their students. Studies have also shown that student-teacher interactions affect girls' engagement with STEM. Teachers often give boys more opportunity to figure out the solution to a problem by themselves while telling the girls to follow the rules.
\\
   \cellcolor{blue!30}
\textbf{Passage 2\quad Source: The New York Times\quad Stance: -24.01} \\   
    \cellcolor{blue!10}
    How are the these pressures different for girls and women than they are for boys and men? If you could change one thing about typical and/or stereotypical gender roles, what would it be? 2. As a class, read and discuss the article ``Girls Will be Girls'' , focusing on the following questions: a. What does the author, Peggy Orenstein, mean when she says that many women are ``struggling to find an ideal mix of feminism and femininity''? Do you agree? Why or why not? b. Why did some people get upset about the implicit ``Girls Keep Out'' sign on the cover of the ``Dangerous Book for Boys''? \\
   \cellcolor{blue!30}
\textbf{Passage 3\quad Source: The New York Times\quad Stance: -7.91} \\   
    \cellcolor{blue!10}
    Poverty is becoming defeminized because the working conditions of many men are becoming more feminized. Whether they realize it or not, men now have a direct stake in policies that advance gender equity. Most of the wage gap between women and men is no longer a result of blatant male favoritism in pay and promotion. Much of it stems from general wage inequality in society at large. IN most countries, women tend to be concentrated in lower-wage jobs. The United States actually has a higher proportion of skilled and highly paid female workers than countries like Sweden and Norway. Yet as a whole, Swedish and Norwegian women earn a higher proportion of the average male wage than American women because the gap between high and low wages is much smaller in those countries. \\
   \cellcolor{blue!30}
\textbf{Passage 4\quad Source: The New York Times\quad Stance: -21.75} \\   
    \cellcolor{blue!10}
    Site Navigation Site Mobile Navigation Women and the Pay Gap In ``How to Attack the Gender Wage Gap? Speak Up'' (Dec. 16), a solution is proposed for the problem of pay inequality: make women stronger negotiators in securing their own salaries. But we should always remember that employers have an obligation to follow the law in the first place, and to pay men and women working in the same jobs the same pay. Fifty years ago, Congress decided as much by passing the Equal Pay Act - but since then the wage gap has narrowed little. There's nothing wrong in women honing their negotiating skills - and some will succeed in getting higher pay.
    \\
   \cellcolor{yellow!50}
\textbf{Passage 5\quad Source: The Washington Post\quad Stance: -6.89} \\   
   \cellcolor{yellow!20}
    But, under this metric for people with a college degree, there is virtually no pay gap at all.'') To be specific, ``The [Bureau of Labor Standards] reports that single women who have never married earned 96\% of men's earnings in 2012,'' two conservative economists relate. (The Post's Wonkblog comes up with 91 percent. ) Job selection, risk taking (``Nearly all the most dangerous occupations, such as loggers or iron workers, are majority male and 92\% of work-related deaths in 2012 were to men. Dangerous jobs tend to pay higher salaries to attract workers''), the choice to work part-time and factors other than discrimination explain the differences: While the BLS [Bureau of Labor Statistics]reports that full-time female workers earned 81\% of full-time males, that is very different than saying that women earned 8\% of what men earned for doing the same jobs, while working the same hours, with the same level of risk, with the same educational background and the same years of continuous, uninterrupted work experience, and assuming no gender differences in family roles like child care.
    \\
    \end{tabular}
    \vspace{2mm}
    \caption{
    Top 5 retrieved passages after ranking for example in Figure 5 in the main paper.} 
    \label{fig:sample-reranked-1}
    \egroup
\end{figure*}

\begin{figure*}[th]
    \bgroup
    \def\arraystretch{1.5}
    \fontsize{10}{11}\selectfont
    \hspace{-2mm}
    \centering
    \setlength{\tabcolsep}{0.8mm}
    
    \begin{tabular}{|p{140mm}|}
    \hline

    \textbf{Input}: 
    all drugs should be decriminalized. I am a firm believer that MDMA and LSD can be very therapeutic and eye opening. They can expand your consciousness and allow you to see things from a different perspective and have the ability to alter your life. (...) \\
    \hline
    \textbf{Human}: If harder drugs like cocaine, meth, heroin were legalized and easy to obtain, people like me would be curious and try them, not knowing that they definitely can ruin your life. Do you honestly believe someone can be a responsible user of legal heroin?
       \\
    \textbf{Seq2Seq}:  
    i do n't think it 's a bad idea . there are plenty of people who are addicted to drugs . i do n't want to be able to do drugs . however , i do n't really care about anyone else . i 'm not going to try to change your view , but i 'm not sure what you 're talking about .
    \\
    \textbf{Seq2seqAug}:
    as far as i can tell , drug use is the source of the use of drug use . in fact , there is a lot of evidence to support that . drug use can be used as a way to reduce crime . the problem is that drug use does n't make sense . it 's like saying " if you do n't like it , it 's illegal . "\\
    \textbf{CANDELA}: there is a lot of research that suggests that there is a substantial amount of drugs that can be used to cause harm to others . however , there is a significant amount of research that can lead to a lot of drug addiction in the us . there are a lot of things that can be attributed to drug use , drug use , substance abuse , etc . i do n't think it 's fair to say that drug use should be illegal , but i think it 's important to note that it 's not a gateway drug . it 's just a symptom of the root cause of the drug addiction problem . it 's not just about the 
    \\
 \textbf{CANDELA \textit{input w/o passages}}: if you want to look at the effects of drug use , you 'll find that there are a lot of people who are addicted to drugs like meth , heroin , and so on . in fact , there are some studies that show that there is a correlation between drug use and the availability of drugs in the us . in other words , there is a significant correlation between the effects of marijuana and drug use , and the correlation between marijuana and marijuana use . this is not to say that drug use is bad , but it is also a symptom of drug use . there are many studies that show a correlation between marijuana use and marijuana are more likely to result in a decrease in violent crime . this is a study that shows that it is not the case that the study found out that it was a result of drug use and it was found that it was found in the study . \\
    \hline
    \end{tabular}
    \vspace{2mm}
    \caption{
    Sample outputs for human and different systems.}
    \label{fig:sample-outputs-2}
    \egroup
\end{figure*}

\begin{figure*}[th]
    \bgroup
    \def\arraystretch{1.5}
	\fontsize{10}{11}\selectfont
    \hspace{-2mm}
    \centering
	\setlength{\tabcolsep}{0.8mm}
	\begin{tabular}{p{140mm}}
	\textbf{Input}: 
    all drugs should be decriminalized. I am a firm believer that MDMA and LSD can be very therapeutic and eye opening. They can expand your consciousness and allow you to see things from a different perspective and have the ability to alter your life. (...) \\
    \hline
    \cellcolor{green!30}
    \textbf{Passage 1\quad Source: Wikipedia\quad Stance: -28.88} \\
    \cellcolor{green!10}
    In 2010, DrugScience produced a ranking of drug harms in the UK, the results of which garnered significant media attention. Drugs were assessed on two metrics - harm to users and harms to society. The report found heroin, crack cocaine, and methamphetamine to be the most harmful drugs to individuals, with alcohol, heroin, and crack cocaine as the most harmful to others. Overall, alcohol was the most harmful drug, with heroin and crack cocaine in second and third places. Most notably, it found the legal status of most drugs bears little relation to the harms associated with them - several class A drugs including ecstacy (MDMA), LSD and magic mushrooms featured at the very bottom of the list.The report found heroin, crack cocaine, and methamphetamine to be the most harmful drugs to individuals, with alcohol, heroin, and crack cocaine as the most harmful to others. Overall, alcohol was the most harmful drug, with heroin and crack cocaine in second and third places. Most notably, it found the legal status of most drugs bears little relation to the harms associated with them - several class A drugs including ecstacy (MDMA), LSD and magic mushrooms featured at the very bottom of the list. Similar findings were found by a Europe-wide study conducted by 40 drug experts in 2015.
\\
  \cellcolor{pink!100}
    \textbf{Passage 2\quad Source: The Wall Street Journal\quad Stance: -5.89} \\
    \cellcolor{pink!50}
    Something drastic needs to be done, and the steps suggested by Mr. Murdoch may be a good start. 3:13 pm October 8, 2010 Anonymous wrote: SHOW ME THE MONEY! 12:09 pm October 11, 2010 JustFacts wrote: Where is the "accountability" for the CIA and other corrupt govt.  Wall Street-affiliated players involved with international drug smuggling for decades (!) -- deliberately inundating communities \& specific neighborhoods with heroin, cocaine, meth, pills (MDMA/ecstacy), etc. It is a documented fact that the CIA \& corrupt elements of the U.S. govt. \& freemasons have been involved in large-scale heroin distribution operations and also involved in the deliberately induced crack cocaine epidemic targeting black neighborhoods (for the purposes of social undermining \& political-economic control). \\
      \cellcolor{yellow!50}
    \textbf{Passage 3\quad Source: The Washington Post\quad Stance: -10.05} \\
   \cellcolor{yellow!20}
    (Photo by Brian Vastag) In the last few years, he saw a resurgence in legitimate research on MDMA as academics restarted clinical trials with MDMA as a therapeutic tool, publishing studies showing that the drug can help veterans come to terms with the trauma of war. ``He was very depressed once MDMA was criminalized,'' said Rick Doblin, president of the Multidisciplinary Association for Psychedelic Studies, which funds clinical trials of MDMA and LSD. ``Sasha always felt these drugs didn't open people up to drug experiences, but opened us up to human experiences of ourselves.'' In 1985, with first lady Nancy Reagan's ``Just Say No'' campaign in full swing, federal authorities banned MDMA with an unusual emergency action. \\
    \cellcolor{yellow!50}
    \textbf{Passage 4\quad Source: The Washington Post\quad Stance: -7.96} \\
    \cellcolor{yellow!20}
    It wasn't MDMA after all, but methamphetamine. A new review board quickly signed on to support Mithoefer's study, but the irony of the wasted year wasn't lost on him: The misidentified drug that had been deemed too toxic to evaluate for medical use, the drug that was far more toxic than MDMA, was already a prescription drug. Meanwhile, in the four years the MDMA study lingered between concept and reality, Donna Kilgore had been driven to the brink. She took "every anti-depressant you can name," tried a dozen therapists and an almost equal number of therapeutic approaches. But nothing made that numbness, panic and rage recede.
    \\
    \cellcolor{blue!30}
    \textbf{Passage 5\quad Source: The New York Times\quad Stance: -5.67}   \\
    \cellcolor{blue!10}

    For alcohol, the safety margin is 10 (330 divided by 33 equals 10). In other words, it takes 10 times as much alcohol to kill you as it does to give you a buzz. Note: All such calculations are very rough estimates, and severe toxic reactions can occur at much lower doses depending on the health of the individual. Based on my research, the safety margin of recreational substances as normally used is: 6 for heroin; 10 for alcohol; 15 for cocaine; 16 for MDMA; 20 for codeine; and 1,000 for LSD or marijuana. Cigarettes have little immediate risk of death because most of the very lethal nicotine is destroyed in the smoke.
    \\
    
    \end{tabular}
    \vspace{2mm}
    \caption{
    Top 5 retrieved passages after ranking for example in Figure \ref{fig:sample-outputs-2}.}
    \label{fig:sample-reranked-2}
    \egroup
\end{figure*}
    
\begin{figure*}[th]
    \bgroup
    \def\arraystretch{1.5}
    \fontsize{10}{11}\selectfont
    \hspace{-2mm}
    \centering
    \setlength{\tabcolsep}{0.8mm}
    \begin{tabular}{|p{140mm}|}
    \hline
    \textbf{Input}: 
    the presidents new plan to restrict welfare beneficiaries from obtaining a green card is a bad policy . (\dots)  legal immigrants who receive any significant form of Welfare or public benefits are given a large "black mark" on their record that is heavily factored in should they attempt to apply for a green card. (\dots) \\
    \hline
    \textbf{Human}:
    The policy is impacting individuals, not the group. Each individual is judged based on the merit they bring forward. If an individual takes government assistance, they are most likely not contributing positively to the US.
    \\ 
     \textbf{Seq2Seq}: 
    i do n't think it 's fair to say that the government should be able to do this . if you want to get rid of illegal immigration , you need to be able to find a way to do it . you need to make sure that you do n't want to get a job . it 's just a matter of if you do n't have a problem with the government . 
    \\
    \textbf{Seq2seqAug}
    there are a lot of people who do n't want to have children , but they do n't have to be able to pay for it . it 's the government 's job to make sure that the immigrants do n't have a job . they do n't need to be deported . they have to go through the process of becoming a citizen . 
    \\ 
   \textbf{CANDELA}: i 'm not going to try to change your view , but i 'd like to point out that there are a number of undocumented immigrants in the united states , and i 'm not sure what you 're talking about , but i do n't think it 's fair to say that undocumented immigrants are going to be more likely to be able to get the benefits of citizenship . i 'm not a fan of this , but i 'm going to focus on the fact that there is a significant amount of people who do n't have access to welfare programs in the us , and that 's not the case . it 's not like 
    \\ 
 \textbf{CANDELA \textit{input w/o passages}}: 
    a lot of illegal immigrants in the united states are not in the same position as they are in the us . in fact , the majority of illegal immigrants do not want to pay for the services provided by the federal government . it 's not that they are undocumented , it 's that they do n't have to worry about the fact that they 're undocumented . they 're not going to be able to get a job because they ca n't afford to live in a society where they ca n't get a job . if you 're talking about illegal immigrants , it 's not like they 're in the same boat as you  \\ 
    \hline
	\end{tabular}
	\vspace{2mm}
	\caption{
     Sample outputs for human and different systems.}
    \label{fig:sample-outputs-3}
    \egroup
\end{figure*}


\begin{figure*}[th]
    \bgroup
    \def\arraystretch{1.5}
	\fontsize{10}{11}\selectfont
    \hspace{-2mm}
    \centering
	\setlength{\tabcolsep}{0.8mm}
	\begin{tabular}{p{140mm}}

	\textbf{Input}: 
    the presidents new plan to restrict welfare beneficiaries from obtaining a green card is a bad policy . (\dots)  legal immigrants who receive any significant form of Welfare or public benefits are given a large "black mark" on their record that is heavily factored in should they attempt to apply for a green card. (\dots) \\
    \hline
    \cellcolor{pink!100}
    \textbf{Passage 1\quad Source:  The Wall Street Journal\quad Stance: -6.57} \\
    \cellcolor{pink!50}
    Who wrote this? Is that you Obama? 11:00 pm February 2, 2011 Welfare Worker in Washington State wrote: I think it is erroneous to assume that a large percentage of welfare recipients are people of color. I see more white people with their hands out in our area. We do have a large percentage of illegals with US born children and immigrants from Russia here in Washington that are receiving benefits. The Russian immigrants bring in their parents and extended family that get SSI benefits for the first 5 years if they are 65 or older. They have large families - and even if they are working have a tendency to get close to 1000 in food benefits each month.
\\
 \cellcolor{blue!30}
    \textbf{Passage 2\quad Source: The New York Times\quad Stance: -9.50 } \\
    \cellcolor{blue!10}
    Temporary workers 883 entries in 2015 Employees (and their families) on non-immigrant work visas like H-1B for specialty workers and H-2B for agricultural workers. Fiances of U.S. citizens 669 entries in 2015 Temporary visas for fiances of U.S. citizens and for spouses and children of U.S. citizens or green card holders who have pending immigrant visas. BARRED New Immigrants Like the original order, the new ban also applies to people from the six countries newly arriving on immigrant visas, which are issued based on employment or family status. People issued immigrant visas become legal permanent residents on arrival in the United States and are issued a green card soon after. \\
     \cellcolor{pink!100}
    \textbf{Passage 3\quad Source:  The Wall Street Journal\quad Stance: -12.09} \\
    \cellcolor{pink!50}
   The family of the Boston bombers (although here legally as "refugees") collected significant amounts of food stamps, housing subsidies, college subsidies and the like. At the same time, they had sufficient funds to travel to their homeland and other European destinations, despite having reported only modest earnings. Last week, it was reported that a Pakistani owner (presumably a legal immigrant) of a chain of 7-11's was importing illegals to work in his stores. We can have a welfare society or open borders but we can't have both. Any immigration reform has to stipulate that immigrants cannot receive any taxpayer funded benefits (federal, state or local) until after they have achieved citizenship. \\
   \cellcolor{blue!30}
    \textbf{Passage 4\quad Source: The New York Times\quad Stance: -6.16 } \\
    \cellcolor{blue!10}
    They are not entitled to a passport or a green card because they bypassed the legal mechanisms for obtaining such documents. In any other country they would be promptly deported, justifiably. I support immigration reform and personally do not feel any economic competition from illegal aliens. That being said, there is a difference between immigrants who have applied for and received citizeship or green cards and those who have not. There should be a fast track naturalization system for children of illegals, such as this student. He grew up here because of his parents' actions, not his own. This is similar to being born here, which has historically entailed citizenship.
    \\
    \cellcolor{yellow!50}
      \textbf{Passage 5\quad Source: The Washington Post\quad Stance: -13.81 } \\
    \cellcolor{yellow!20}
    In 1996, Congress enacted a requirement that legal immigrants be present for five years before becoming eligible for benefits. But we have never categorically excluded immigrants from receiving public benefits. Until now. If approved, the new policy would effectively deter legal immigrants from using public benefits for which they are eligible, lest they later be denied a green card or be removed. The DHS could also apply ``public charge'' to legal immigrants who use benefits for their children (such as CHIP), even if the children are U.S. citizens. The Migration Policy Institute estimates that the new policy could have a chilling effect on some 18 million noncitizens and 9 million U.S.-citizen children who reside in families where at least one person uses Medicaid/CHIP, welfare, food stamps or SSI.
    \\
    
    \end{tabular}
    \vspace{2mm}
    \caption{
    Top 5 retrieved passages after ranking for example in Figure \ref{fig:sample-outputs-3}.}
    \label{fig:sample-reranked-3}
    \egroup
\end{figure*}